\documentclass[letterpaper, 11pt]{IEEEtran} 
\usepackage{graphicx} 
\usepackage[square, numbers]{natbib}  
\usepackage{algorithm}
\usepackage{algorithmic}
\usepackage[utf8]{inputenc} 
\usepackage[T1]{fontenc}    
\usepackage{booktabs}       
\usepackage{amsfonts}       
\usepackage{amsmath}       
\usepackage{nicefrac}       
\usepackage{microtype}      
\usepackage{xcolor}         
\usepackage{subfig}
\usepackage{multirow}
\usepackage{array}
\usepackage{newfloat}
\floatstyle{ruled}
\usepackage{titlesec}
\titleformat*{\section}{\Large\bfseries}
\titleformat*{\subsection}{\large\bfseries}

\usepackage{hyperref}

\author{
    Michael Hobley, Victor Prisacariu \\ Active Vision Laboratory\\
  University of Oxford\\
  \texttt{[mahobley, victor]@robots.ox.ac.uk} \\
}
\title{Learning to Count Anything: Reference-less \\ Class-agnostic Counting with Weak Supervision}

\begin{document}
\urlstyle{tt}
\maketitle

\begin{abstract}
\looseness=-1
Current class-agnostic counting methods can generalise to unseen classes but usually require reference images to define the type of object to be counted, as well as instance annotations during training. Reference-less class-agnostic counting is an emerging field that identifies counting as, at its core, a repetition-recognition task. Such methods facilitate counting on a changing set composition. We show that a general feature space with global context can enumerate instances in an image without a prior on the object type present. Specifically, we demonstrate that regression from vision transformer features without point-level supervision or reference images is superior to other reference-less methods and is competitive with methods that use reference images. We show this on the current standard few-shot counting dataset FSC-147. We also propose an improved dataset, FSC-133, which removes errors, ambiguities, and repeated images from FSC-147 and demonstrate similar performance on it. To the best of our knowledge, we are the first weakly-supervised reference-less class-agnostic counting method. 
\end{abstract}

\section{Introduction}
\label{Introduction}

Counting is one of the first abstract tasks people learn. Once learnt, the concept is simple. Its aim is to find the number of instances of an object class. 
    Simple though it is, counting has diverse applications including: crowd-counting, traffic-monitoring, conservation, microscopy, and inventory management.
    Significantly, whereas people can generally count objects without a prior understanding of the type of the object to be counted, most current automated methods cannot.
    
    \looseness=-1 Presented with a set of novel objects and asked to `count', a person would know what is to be counted. This does not require a reference example or prior understanding of object type to clarify that we do not want to find the repetitions of a self-similar background.
    The ability to count is thus at its core comprised of two components: an understanding of what could be worth counting and an ability to identify repetitions of those countables.
    We demonstrate that self-supervised vision transformer features, specifically their use of self-attention with a global receptive field, are crucial for both conditions to be met. 
    We prove experimentally that, given general and globally contextual features, enumerating instances is simple and  requires only minimal training and supervision. 
    
    \begin{figure}[t]
    \centering
    \includegraphics[width=\linewidth]{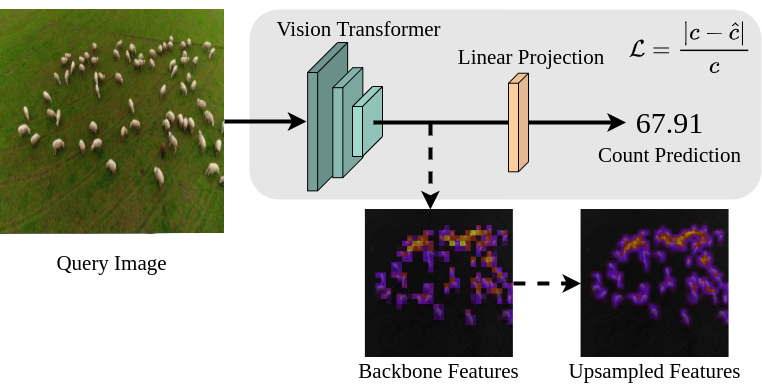}
        \caption{
    \textbf{The RCC pipeline.} 
    Our method learns to count objects of novel classes without reference images, using only the ground truth count, $c$, as supervision.
    We also visualise these features to show they suffice to  localise instances of the counted object.
    }
    \label{pipeline}
\end{figure}
    
    Previous methods, whether detection-based, regression-based, or classification-based, generally focus on enumerating the instances of a single or small set of known classes, such as people \cite{wang2015deep, cao2018scale},
    vehicles \cite{mundhenk2016large},
    animals \cite{go2021fine}, 
    or
    cells \cite{xie2018microscopy}.
    This requires an individually trained network for each type of object with limited to no capacity to adapt to novel classes.
    Individually trained networks require gathering new data and retraining whenever a new type is considered, which is difficult and expensive.  
    Additionally, these methods generally aim to localise instances before enumerating them, requiring point-level annotations to supervise the training. 
    These class-specific and point-level supervised systems are only feasible when the composition and appearance of object types will remain the same indefinitely and point-level annotations exist, which is often not the case in real-world applications. 
    In contrast, class-agnostic counting methods \cite{Lu18, ranjan2021Famnet} do not require a static class composition as they adapt an understanding of counting learnt on a set of known classes to objects of unseen classes.
    However, most class-agnostic methods still require training-time point-level annotations and reference images of the class to count. The only exception is RepRPN \cite{ranjan2022exemplar}, a twostage method that does not need reference images but uses point-level annotations.
    Weakly-supervised counting methods \cite{lei2021towards, yang2020weakly} relax the need for point-level annotations but do not currently generalise to unseen classes.

    Given the cost of gathering reference images and point annotations in dynamic, real-world applications, we present a method that accurately enumerates instances of an unseen class without using such images or annotations. Our simple method achieves this by creating a general, global-context aware feature space using a vision transformer that can then be enumerated with a linear projection, see Figure \ref{pipeline}.

    Our key contributions are as follows:
    \begin{enumerate}
        \item We propose RCC, a reference-less class-agnostic counter, and show
        it can be trained without instance annotations.
        \item We demonstrate that RCC outperforms the only other reference-less class-agnostic counting method and is competitive with current methods that use reference images as a prior and train with full point-level supervision. 
        \item We present FSC-133, an improved dataset for class agnostic counting without the errors, ambiguities, and repeated images present in FSC-147.
    \end{enumerate}
\section{Related Work}
\label{relatedwork}

\textbf{Class-specific counting.}
\looseness=-1
Class-specific counting methods enumerate instances of known object types in an image. 
These approaches can be broadly grouped into detection-based, regression-based, and classification-based methods.
\textbf{Detection-based} methods use standard detection or segmentation approaches \cite{ren2015fasterrcnn, mundhenk2016large, michaelis2018one} to find all objects of a type \cite{hsieh2017drone} or set of types \cite{desai2011discriminative}, and then enumerate them. 
Detection is itself a broad field with significant recent developments that can be beneficial for counting applications, such as in handling overlap or occlusion. However, detection-based counting methods are still unsatisfactory in high-density applications.
\textbf{Classification-based} methods \cite{mundhenk2016large} generate a discrete classification of an image's global count.
Their most obvious flaw is that they treat all incorrect counts, independent of proximity to the ground-truth count, equally. This makes training difficult, needs large amounts of data, and requires a low maximum count to ensure each discrete count-class is correctly trained.
\textbf{Regression-based} counting methods aim to regress a single global count \cite{chan2009bayesian, wang2011automatic, wang2015deep} or a pixel-level density map prediction, which can be enumerated by integration \cite{cao2018scale, ranjan2021Famnet, xie2018microscopy} or instance detection \cite{barinova2012detection, arteta2014interactive, cholakkal2020towards}. These can use complex features \cite{zhang2015cross,zhang2016single} or simpler low-level features \cite{chan2009bayesian}, such as texture \cite{marana1997estimation}. 
A regressed density map can also be used as a rudimentary object detector, followed by cross-correlation with a reference to find instances of the desired class \cite{sokhandan2020few}.

\textbf{Weakly-supervised counting.} While the majority of counting methods above use some form of instance-wise positional information, gathering this information is costly. 
Weakly-supervised counting methods aim to generate an accurate count with minimal \cite{lei2021towards, sam2019almost} or no point-level annotations \cite{borstel2016gaussian, yang2020weakly, wang2015deep}.

\textbf{Class-agnostic counting.}
The above methods assume prior understanding of all object classes to be identified. 
Generally, this requires an individually trained system for each class of objects, with limited capacity to adapt to novel classes. 
To avoid the cost of retraining these non-generalisable methods to new classes, \citet{Lu18} proposed class-agnostic counting, a framework in which test-time classes are not present during training.
However, this and most subsequent class-agnostic methods \cite{ranjan2021Famnet, yang2021cfoc,sokhandan2020few} require a prior of object class at test time, in the form of reference images. The exception is a concurrent work  \cite{ranjan2022exemplar}, which uses a two-step process which proposes regions likely to contain an object of interest, and then uses these regions for a reference-based method. 
Class-agnostic counting has so far been achieved by creating a sufficiently general feature space and applying some form of matching to the whole feature map \cite{yang2021cfoc, shi2022represent} or to proposed regions of interest \cite{ranjan2021Famnet, sokhandan2020few}.

\textbf{Contextual features and vision transformers.}
\looseness=-1
Convolutional neural networks, which most of the above methods use, fail to accurately understand global context due to their localised receptive fields. 
Methods have tried to solve this using dilation \cite{li2018csrnet, bai2020adaptive} or rank-based \cite{liu2018leveraging} systems. 
Attention-based architectures that model global context inherently have been used to generate more contextually-aware features to aid counting \cite{sindagi2019ha}, to perform feature matching \cite{shi2022represent, lin2021object} and to generate reference image proposals \cite{ranjan2022exemplar}. 
Inspired by vision transformers such as ViT \cite{dosovitskiy2020vit} and DETR \cite{carion2020DETR}, there have been various counting developments \cite{do2021attention,liang2022end, sun2021boosting}, including some \cite{liang2021transcrowd, savner2022crowdformer, wang2022joint} with a specific focus on weakly-supervised counting. These methods, however, focus on crowd-counting, a class specific task with limited generalisability.
\section{Method}
\label{method}
In this paper, we approach the challenging task of counting instances of unseen classes without reference images or point-level supervision. 
We believe that reference-less counting can be broken into two fundamental questions: `Where are object instances we may want to count?' and `Is the class of this given instance repeated elsewhere?'. 
Unlike RepPRN \cite{ranjan2022exemplar}, the only other reference-less method, we solve these questions simultaneously.
The former, in a class-agnostic context, requires a general, informative feature space. We identify that self-supervised knowledge distillation is well-suited to learning such a feature space. The latter requires an understanding of the global context of an image. Vision transformers, and their use of attention, are perfectly suited to this task.
Given a feature space that is general and globally aware, we find that regressing a count is simple and can be achieved with minimal training and a single linear projection without point-level supervision.

\textbf{Self-supervised knowledge distillation.}
In order to learn a general, informative feature space without the use of labels, we use a training methodology of self-supervised knowledge distillation based on \citet{caron2021DINO}. We learn informative representations of images by encouraging consensus between a fixed `teacher' network, $g_{t}$ parameterised by $\theta_{t}$,  which operates on large `global' crops, $x_{1}^{g}$, $x_{2}^{g}$,  of the image and a `student' network, $g_{s}$ parameterised by $\theta_{s}$, which operates on a set, $V$, of smaller, `local' crops of the image.
This is achieved by minimising the cross-entropy between the probability distributions of these networks, $P_{t}$ and $P_{s}$, as in Equation \ref{eq_cross_entropy}.
These probability distributions, defined in Equation \ref{eq_dino_predictions}, act as `unsupervised classification predictions'. 
\begin{equation}
\label{eq_cross_entropy}
    \min\limits_{\theta_{s}}
    \sum_{x \in \{ x_{1}^{g}, x_{2}^{g}\}}
    \; \;
    \sum\limits_{\substack{x' \in V \\ x' \neq x}}  H(P_{t}(x), P_{s}(x'))
\end{equation}
\begin{equation}
\label{eq_dino_predictions}
    P_{s}(x)^{(i)} =
    \frac{\exp(g_{s}(x)^{(i)}/\tau_{s})}{\sum_{d=0}^{d_p}\exp(g_{s}(x)^{(d)}/\tau_{s})}
\end{equation}

\noindent 
Here, $H(a,b) =-a\log b$, $d_p$ is the dimensionality of the probability distribution, and
$\tau_{s} > 0$ is a temperature parameter that controls the sharpness of the distribution. 
A similar formula to Equation \ref{eq_dino_predictions} holds for $P_{t}$ with temperature $\tau_{t}$.
We iteratively update the weights of the teacher network using an exponential moving average of the student network's weights.

\textbf{Vision transformer backbone.}
The crucial global context afforded by transformers stems from their use of an attention mechanism. Attention generates a new feature from linear projections of all other features based on their similarity.
Inspired by \citet{vaswani2017attention}, our vision transformers, $g_{t}$ and $g_{s}$, use multiple attention heads to generate informative, diverse features, as in Equation \ref{eq_multihead} and Equation \ref{eq_multihead_2}.
\begin{equation}
\label{eq_multihead}
    g_s(x; \theta_s) = \text{Concat}(head_1, ..., head_h)
\end{equation}
\begin{equation}
\label{eq_multihead_2}
    head_i =  \text{Attention}(q_i,k_i,v_i; \theta_{s})
\end{equation}
where $q_i$, $k_i$, and $v_i$, the queries, keys, and values, are each a linear projection of a patch of $x$. 
Both $q_i$ and $k_i$ have the dimensionality $d_k$. In the case of self-attention, $q_i = k_i$.
In practice, each head's attention is calculated simultaneously for $Q_i$, $K_i$, and $V_i$, sets of $q_i$, $k_i$, and $v_i$ respectively, as:
\begin{equation}
\label{eq_attention}
    \text{Attention}(Q_i,K_i,V_i) = \text{softmax}(\frac{Q_iK_i^T}{\sqrt{d_k}})V_i 
\end{equation}

\textbf{Count regression.}
\looseness=-1
We directly regress a single, scalar count prediction
for the whole input image, $x$, from the latent features of $g_s(x)$ without point-level annotations, which are difficult and expensive to gather.
Although it may seem that a location-based loss should aid in training, we found that point-level annotations and Gaussian density maps were unnecessary and often detrimental in a class-agnostic setting. 
Since positional annotations are often arbitrarily placed and contain at most limited information about the size and shape of an object, identifying different parts of all correct objects would be punished by a positional loss function. This arbitrary punishment hinders the network's understanding of the counting task.
We allow the network to develop its own conceptual representation of the task by regressing an estimate for the count directly. To this end, we use one of the simplest possible loss functions, the absolute percentage error, defined as: $\mathcal{L} = \text{Absolute Percentage Error}= \lvert c - \hat{c} \rvert/c $, where $c$ is the ground truth count and the predicted count $\hat{c} = F \cdot g_s(x)$, where $F$ is a learnt linear projection. This was a slight improvement over Absolute Error as it limited the disproportionate effect of very high-density images on the gradients.
We found that the network, without the imposition of human ideas of positional salience, learnt to implicitly localise instances of the counted class in a meaningful way, as further discussed in Section \ref{results_visualisation} and seen in Figure \ref{figure_visualisation}.

\textbf{Tiling augmentation.}]
\looseness=-1
While multi-scale systems have been used to improve detection for objects of varying size, they require a complex spatial loss \cite{redmon2016YOLO} or non-maximal suppression \cite{ren2015fasterrcnn}, which has a large computational cost.
As our method regresses a single count rather than a set of object locations, these approaches are not feasible.
To allow the network to better understand  multiple scales and densities, we instead increase the diversity of image densities at training time by tiling resized versions of the input image into a (2$\times$2) grid for 50\% of instances.  This increased the representation of high-density images.
This augmentation improved our methods MAE and RMSE by about 10\% and 20\% respectively. See Appendix \ref{append_tile} for details.

\section{FSC-147 and FSC-133}
FSC-147 \cite{ranjan2021Famnet} is a recent dataset widely used in the field of class-agnostic counting. 
It is meant to contain 6135 unique images from 147 distinct classes. To evaluate the generalisability of a method to unseen classes, the classes and images for training, validation, and testing are not meant to  overlap.
Although our method does not use them, this dataset also includes point annotations for each instance and three random instance bounding box annotations per image. 

We found 159 images that appear 334 times in the dataset with a pixel-wise difference of 0 with at least one other image when compared at a $224 \times 224$ resolution.
If we include images that are close to identical but do not have a zero pixel-wise difference, these numbers increase to 211 images that appear 448 times.
See Appendix  \ref{similaranddifferent}, \ref{list_identical} for illustration 
and the complete lists of duplicates.
Further, 11 images appear in the training set and one of the validation or testing sets.
This significant issue undermines the purpose of these splits. In addition, in 71 instances, an image repeats with different counts,  with discrepancies of up to 25\%, including 5 which appear in both the training and the validation sets with 9\%-21\% discrepancy. See Appendix  \ref{discrepancies} for the full list.

\looseness=-1
We, therefore, propose a revised version of this dataset, FSC-133, which corrects for these issues.
The data split overlap stems from misclassification. This is likely due  to ambiguous class distinctions (eg. `kidney beans' and `red beans') or hierarchical classes (eg. `cranes', `seagulls', and `geese' could all also be classified as `birds'). In both cases, it can be difficult or even arbitrary to identify which class is most appropriate. 
`Cranes' and `geese' or `bread rolls', `buns', and `baguette rolls' are similar enough that one might well classify instances of each as of the type with which they are most familiar. See Figure \ref{arethesame} in Appendix  \ref{similaranddifferent} for visual examples of unclear class divisions.
To disambiguate the dataset and help it more reliably measure a method's generalisability, FSC-133 combines categories that are similar enough to be easily confused, see Appendix  \ref{append_classes} for details.
The count discrepancies seem to occur when there is occlusion between objects or partial objects appear at the edges of the image, see Appendix  \ref{similaranddifferent} for examples. 
While these images are difficult for a human to accurately count, we believe that they are of value to the dataset. Instead of removing them, FSC-133 includes only the most accurate count.

FSC-133 has 5898 images in 133 classes. The training, validation, and testing sets have 3877, 954, and 1067 images from 82, 26, and 25 classes respectively.
When combining classes that overlapped data splits, we merged them into the training split so that methods trained on FSC-147 be tested on FCS-133. While disadvantaged, these tests would still fairly evaluate the method's ability on completely unseen classes.

\section{Experiments}
\label{experiments}
In this section, we discuss the details of our implementation, training, and the metrics we use for evaluation.
\subsection{Architecture}
\label{experiments_implementationdetails}
We use a ViT-small backbone inspired by DeiT-S \cite{touvron2021training}. This was chosen due to its efficiency and to provide a good comparison to other counting methods.
ViT-S has a similar number of parameters (21M vs 23M), throughput (1237im/sec vs 1007im/sec), and supervised ImageNet performance (79.3\% vs 79.8\%) as ResNet-50 \cite{touvron2021training}, which is used by contemporary counting methods.

As even data efficient transformers require relatively large amounts of data to train and the datasets on which we evaluate are small, we initialised our transformer backbone, $g_s$, using weights from \citet{caron2021DINO}. This self-supervised pre-training gives the network an understanding of meaningful image features without supervision and prior to exposure to our limited datasets, minimising the chance of overfitting. 
Our linear count projection $F$, projects from a $p\times d_m$ feature space to a scalar count prediction,
where $d_m$ is the dimensionality of the transformer features and $p$ is the number of patches of the vision transformer. We found $d_m = 384$ and $p = 28^2$ sufficient to achieve competitive results while also being lightweight enough to be trainable on a single 1080Ti. It should be noted that this vision transformer configuration limits the resolution of our input image to (224$\times$224) as opposed to the $\sim$(384$\times$384) used by contemporary methods with ResNet-50 backbones.
The code to reproduce our results will be made publicly available at: \url{https://github.com/ActiveVisionLab/LearningToCountAnything}.

\subsection{Training}
\label{experiments_training}
We used a batch size of 2, distributed over 2 GPUs (Titan X). We trained for 80 epochs with a learning rate of $3\mathrm{e}-5$, which takes 2.5 hours.
To increase the diversity of object densities 
at training time, we applied our 2 × 2 tiling augmentation to 50\% of the iterations.
We applied random reflections and rotations to the images and, when tiled, to each tile independently. Colour-based augmentations (colour-jitter, Gaussian blur, and solarisation) had negligible effect on our results. We did not apply random crops as this would require point-level annotations to adjust the count. 
\subsection{Evaluation Metrics and Trivial Baselines}
\label{experiments_evaluationmetrics}
In accordance with previous works on class-agnostic counting \cite{shi2022represent, ranjan2022exemplar}, we use Mean Absolute Error ($\text{MAE} = (\sum_{i=1}^{n_{test}} |c_i - \hat{c}_i|)/n_{test}  $) and Root Mean Squared Error ($\text{RMSE} = \sqrt{(\sum_{i=1}^{n_{test}} (c_i - \hat{c}_i)^2)/n_{test}}$) to evaluate our performance, 
where $c_i$ and $\hat{c}_i$ are the ground truth and predicted count for image $x_i$, and $n_{test}$ is the number of images in the validation or test set.

We compare our method and previous methods to two trivial baselines. Both predict the same value, $\hat{c}$, for all test images, as follows: $\hat{c}_{mean} =(\sum C_{train})/n_{train}$ and $\hat{c}_{median} = C_{train}[n_{train}/2]$, 
where $C_{train}$ is an ordered list of all ground truth counts for the training set and $n_{train}$ is the number of images in the training set.
\section{Results}
In this section, we show that RCC dramatically outperforms other reference-less class-agnostic counting methods and is competitive with reference-based methods. Since our method does not need reference images or point-level annotations, it can be applied to broader real-world applications where the objects to count are ever-changing. We validate this by showing our method's ability to generalise to a novel domain. 
We also use our learnt latent features to localise instances in an image as this may have real-world utility and to substantiate that RCC uses meaningful information to count. We then discuss the failure cases and limitations of our method to inform and motivate possible future research. We finally validate specific components of our method.
\subsection{Benchmarking Methods}
We evaluate our method against two trivial baseline methods, two few-shot detection methods, FR \cite{kang2019few} and FSOD \cite{fan2020few},  six reference-based class-agnostic counting methods, GMN \cite{Lu18}, MAML \cite{finn2017model}, FamNet \cite{ranjan2021Famnet}, CFOCNet  \cite{yang2021cfoc}, BMNet \cite{shi2022represent}, LaoNet \cite{lin2021object}, and the only other  class-agnostic counting method which doesnt require a reference image, RepRPN-Counter \cite{ranjan2022exemplar}, see Table \ref{FSC147_results_table}. We also include reference-less modifications to reference-based methods implemented by \citet{ranjan2022exemplar}, denoted by a *. 
Since \citet{ranjan2022exemplar} have not yet released the implementations of their method and of the modified reference-based methods they evaluate, the experiments we could run comparing to them were limited.
To ensure fair comparison, we also train recent, high performing methods with released implementations using the same vision transformer backbone, ViT-S, as our method, denoted by $\dag$ in the results tables.
\subsection{FSC-147 and FSC-133.}
We achieve significantly better result than RepRPN and all reference-less versions of reference-based methods on FSC-147 without the need for point-level annotations. We are also competitive with previous few-shot or reference-based methods on FSC-147 and FSC-133 without the need for reference images, point-level annotations, or test-time adaptation, see Tables \ref{FSC147_results_table} and \ref{FSC133_results_table}. 
This result demonstrates that the combination of an architecture well-suited for counting and a simple, count-centred objective function can learn a meaningful conceptual representation of counting without any location-based information and so does not require reference images.
As will be discussed in Section \ref{failure_cases}, we believe that the discrepancy in FSC-147 test-set RMSE is due to  poor performance on the high-density images in the test set, as these outliers have a greater effect on RMSE than MAE.

We found that methods that were modified to use  ViT-S in place of their ResNet-50 backbones performed worse;
this is likely due to the significant decrease in input image resolution.
Since FamNet was previously using features from two sequential layers of a ResNet backbone, this modification also halved the dimensionality of the features used to regress the count, decreasing their scale invariant performance.

In general, the methods perform better on FSC-133 than FSC-147 with the exception of the validation MAE. The greater validation MAE is likely due to the removal of duplicate images and similar classes. The other metric improvements are likely due to a few high-density images being moved to the training set.
\begin{table}
    \centering
    \fontsize{9}{9}\selectfont
    
\begin{tabular}{lrrrr}
 \toprule
 & \multicolumn{2}{c}{Val Set} & \multicolumn{2}{c}{Test Set} \\
\multicolumn{1}{c}{Method} & \multicolumn{1}{c}{MAE} & \multicolumn{1}{c}{RMSE} & \multicolumn{1}{c}{MAE} & \multicolumn{1}{c}{RMSE} \\

\midrule
Mean & 53.38 & 124.53 & 47.55 & 147.67 \\
Median & 48.68 & 129.7 & 47.73 & 152.46 \\
\midrule
\textit{Reference-based} & & & & \\
MAML 
\cite{finn2017model}
& 25.54 & 79.44 & 24.90 & 112.68 \\
FR 
\cite{kang2019few}
& 45.45 & 112.53 & 41.64 & 141.04 \\
FSOD 
\cite{fan2020few}
& 36.36 & 115.00 & 32.53 & 140.65 \\
GMN (pretrained) 
\cite{Lu18}
& 60.56 & 137.78 & 62.69 & 159.67 \\
GMN 
 \cite{Lu18} \hfill (1-shot) 
& 29.66 & 89.81 & 26.52 & 124.57 \\
FamNet 
 \cite{ranjan2021Famnet} \hfill (1-shot) 
& 26.55 & 77.01 & 26.76 & 110.95 \\
FamNet 
 \cite{ranjan2021Famnet} \hfill (3-shot) 
& 24.32 & 70.94 & 22.56 & 101.54 \\
FamNet+ 
 \cite{ranjan2021Famnet} \hfill (3-shot) 
& 23.75 & 69.07 & 22.08 & 99.54 \\
FamNet$\dag$  \cite{ranjan2021Famnet} \hfill (3-shot) & 37.90 &  109.76 & 33.51  & 137.79  \\
FamNet+$\dag$  \cite{ranjan2021Famnet} \hfill (3-shot) & 37.77 & 109.04 &  33.18 & 137.20
 \\
CFOCNet \cite{yang2021cfoc}
\hfill (3-shot) 
& 21.19 & 61.41 & 22.10 & 112.71  \\
LaoNet \cite{lin2021object}
\hfill (1-shot) 
& 17.11 & \textbf{56.81} & 15.78 & 97.15 \\
BMNet 
  \cite{shi2022represent} \hfill (3-shot) 
& 19.06 & 67.95 & 16.71 & 103.31  \\
BMNet+  \cite{shi2022represent} \hfill (3-shot) 
& \textbf{15.74} & 58.53 & \textbf{14.62} & \textbf{91.83} \\
BMNet$\dag$ 
 \cite{shi2022represent} \hfill (3-shot) 
& 19.29  & 68.58 & 18.34  & 115.31  \\
BMNet+$\dag$  
 \cite{shi2022represent} \hfill (3-shot) 
& 17.21 & 60.18 & 16.90 & 107.66  \\
\midrule
\textit{Reference-less} & & & & \\
MAML* 
\cite{finn2017model}
& 32.44 & 101.08 & 31.47 & 129.31 \\
GMN* 
\cite{Lu18}
& 39.02 & 106.06& 37.86& 141.39 \\
 FamNet*(pretrained) 
 \cite{ranjan2021Famnet}
 & 39.52 & 116.08 & 39.38 & 143.51 \\
FamNet*
\cite{ranjan2021Famnet}
& 32.15 & 98.7 5& 32.27 & 131.46 \\
 RepRPN-Counter 
 \cite{ranjan2022exemplar}
 & 29.24 & 98.11 & 26.66 & 129.11 \\
RCC (ours) & \textbf{17.49}  & \textbf{58.81} & \textbf{17.12}  & \textbf{104.53} \\
\bottomrule
\end{tabular}

    \caption{\textbf{Comparison to state-of-the-art methods on FSC-147.} We outperform other reference-less methods and achieve competitive results with methods which use reference images and test-time adaptation. We highlight the best results for reference-based and reference-less methods. $\dag$ denotes methods trained using the same backbone as ours. 
    * indicates a reference-less modification of a reference-based method.
    }
    \label{FSC147_results_table}
\end{table}
\begin{table}
    \centering
    \fontsize{9}{9}\selectfont
    \begin{tabular}{lrrrr}
 \toprule

 & \multicolumn{2}{c}{Val Set} & \multicolumn{2}{c}{Test Set} \\
\multicolumn{1}{c}{Method} & \multicolumn{1}{c}{MAE} & \multicolumn{1}{c}{RMSE} & \multicolumn{1}{c}{MAE} & \multicolumn{1}{c}{RMSE} \\
\midrule
Mean & 54.39 & 112.68 & 44.76 & 104.28 \\
Median & 51.29 & 119.29 & 43.15 & 109.83 \\
    \midrule

\multicolumn{3}{l}{\textit{Reference-based (3-shot)}} \\
FamNet 
 \cite{ranjan2021Famnet}
& 25.49 & 68.21 & 21.05 & 43.48 \\
FamNet+ 
 \cite{ranjan2021Famnet}
& 24.75 & 67.04
 &  20.20 & 41.76  \\ 
FamNet$\dag$ 
 \cite{ranjan2021Famnet}  &  
 36.89 & 92.76 & 30.09 & 90.95 \\ 
FamNet+$\dag$ 
 \cite{ranjan2021Famnet}
 & 36.36 & 89.22 &  30.79 & 90.71\\ 
BMNet 
 \cite{shi2022represent} & 20.16 & 54.83  & 14.61  & 41.11 \\
BMNet+ 
 \cite{shi2022represent}
&\textbf{16.54} &  \textbf{50.65}  & \textbf{13.85}  & \textbf{40.61}  \\
BMNet$\dag$ 
 \cite{shi2022represent}
& 22.06 & 61.01 & 16.38 & 54.13\\ 
BMNet+$\dag$  
 \cite{shi2022represent}
& 18.78 & 59.76  & 13.87  & 47.27  \\
\midrule
\multicolumn{3}{l}{\textit{Reference-less}} \\
RCC (ours) & \textbf{19.84}  & \textbf{55.81} & \textbf{14.23}  & \textbf{43.83} \\
\bottomrule
\end{tabular}

    \caption{\textbf{Comparison to available state-of-the-art methods on FSC-133.}  We are competitive with reference-based methods without  reference images or test-time adaptation.
    $\dag$ denotes methods trained using the same backbone as ours.
    }
    \label{FSC133_results_table}
\end{table}
 \subsection{Cross-Dataset Generalisability}
To confirm our cross-dataset generalisability, we test our model on CARPK \cite{hsieh2017drone}, a car counting dataset
comprised of birds-eye views of parking lots which significantly differ from the appearance of any objects in FSC-147 or FSC-133. To ensure we are fairly testing genrealisability, we exclude the "car" class from our FSC-133 pretraining. 
We significantly outperform FamNet with and without fine-tuning, and we are competitive with BMNet without fine-tuning. We believe the fine-tuning result discrepancy with BMNet is caused by 
 out-of-distribution high-density images as
70\% of the test images have more instances than the maximum count seen during training.
Both other methods use reference images which  aid in the cross-dataset generalisation and with issues of out of distribution high-density test images. 
Further, while our method improves after fine-tuning, showing the benefit of task-specific information, the improvements are smaller than with FamNet and BMNet. 
This indicates our method's generalisability is relatively optimised.

\begin{table}
    \centering
    \fontsize{9}{9}\selectfont
    \begin{tabular}{lccc}
    \toprule
    Method    & Fine-Tuned & MAE & RMSE   \\ 
    \midrule
    Mean &   N/A  & 65.63 & 72.26 \\
    Median  & N/A  &   67.88 &  74.58 \\
    \midrule
    \multicolumn{3}{l}{\textit{Reference-based}} \\
    FamNet \cite{ranjan2021Famnet}   &   $\times$ &  28.84 & 44.47 \\
    BMNet \cite{shi2022represent}    &   $\times$ &  14.61 & 24.60 \\
    BMNet+  \cite{shi2022represent}  &   $\times$ &  \textbf{10.44} & \textbf{13.77} \\
    FamNet \cite{ranjan2021Famnet}  &  \checkmark &  18.19 & 33.66 \\
    BMNet  \cite{shi2022represent}  & \checkmark  &  8.05 & 9.70 \\
    BMNet+  \cite{shi2022represent}  & \checkmark &  \textbf{5.76}  & \textbf{7.83}  \\
    \midrule
    \multicolumn{3}{l}{\textit{Reference-less}} \\
     RCC (ours)    &  $\times$ &  \textbf{12.31} &
    \textbf{15.40}\\ 
    RCC (ours)    & \checkmark & \textbf{9.21} & \textbf{11.33}\\
    \bottomrule
\end{tabular}

    \caption{\textbf{Generalisation performance on CARPK.} ``Pretrained'' models were trained on FSC-147 or, for our method, FSC-133. ``Fine-tuned'' models were further trained on the CARPK dataset. 
    We highlight the top results with and without reference images and with and without fine-tuning.
    \label{table_carpk}
}
\end{table}

\begin{figure*}
    \centering
    \input{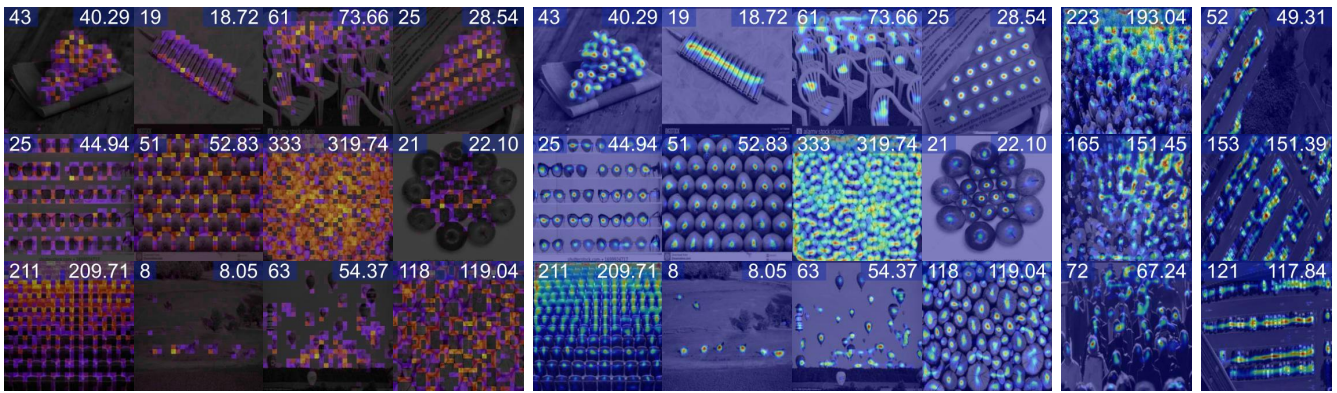}
    \caption{\textbf{Localisation of latent counting features.} Columns 1-4: patch-wise principal components of the counting backbone features, Columns 5-8, 9, 10: pixel-wise localisation predictions of unseen dataset classes from FSC-133, CARPK and ShanghaiTech respectively. The latent counting feature density maps are generated by a localisation head trained on FSC-133.
The count and prediction of each image are in the top left and the top right respectively.  Best viewed in colour and by zooming in.
    }
    \label{figure_visualisation}
\end{figure*}
\subsection{Feature Visualisation}
\label{results_visualisation}
\looseness-1
In order to validate that our network is learning meaningful information, we visualised the most significant value from a singular value decomposition of the counting features weighted by the linear projection weights, see Figure \ref{figure_visualisation}.
As our patch-wise counting features are relatively low resolution and there may be utility in more accurately localising instances in an image, we trained a localisation head. This head, comprised of 3 Conv-ReLU-Upsample blocks, increases the  patch-wise resolution of the trained counting features (28$\times$28) to a  pixel-wise density map prediction (224$\times$224), see Figure \ref{figure_visualisation}. This is trained using the pixel-wise mean squared error of the predicted density map and a ground truth Gaussian density map as is standard in the field \cite{ranjan2021Famnet}. Since this training is not weakly-supervised, we freeze the feature backbone. This training takes 10 epochs and has significantly better results than the same training on a self-supervised backbone, validating our network's ability to count rather than just detect objects. We also show that the backbone and localisation head trained on FSC-133 are generalisable to other domains by testing on the CARPK and ShanghaiTech \cite{zhang2016single} datasets, see Figure \ref{figure_visualisation}.
\subsection{Failure Cases and Limitations}
\label{failure_cases}
\begin{figure}
    \centering
    \fontsize{9}{9}\selectfont
    \input{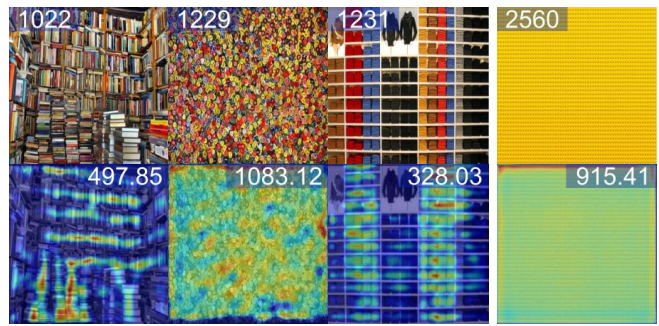}
    \caption{\textbf{High-density images from FSC-133.} The three images from the validation set (left) and the single test image (right) with associated counts of over 1000. 
    }
    \label{FSC147_exclusions_figure}
\end{figure}

A clear failure case of RCC is its difficulty with high-density images.
As shown in Table \ref{FSC147_exclusions_table}, the metrics on FSC-133 improve dramatically when the few images with over 1000 objects are removed.
These images, presented in Figure \ref{FSC147_exclusions_figure}, constitute 0.31\% of the validation set and 0.09\% of the test set.
This failure case is likely because our patch-based approach limits the functional resolution to a $(28\times28)$ feature map.
With many of these high-density images, the same object is found in all patches and across all patch boundaries equally, making individual instances indistinguishable. 
Ideally, we would use a smaller patch size, but due to computational constraints, we could not test this.
As a proxy for this test, we split the high-density images into multiple sub-images that were processed independently, and combined their counts.
This improved the  MAE and RMSE by 45.6\% and 64.9\% respectively. 
However, as this does not generalise in a principled manner to all cases, especially very low density cases, we did not include this as a contribution.

This failure case could also be attributed to the underrepresentation of very high-density images in the training data; only 9 (0.23\%) of 3877 training counts are over 1000. 
This is supported by the fact that these effects appear more dramatically in FSC-147 where high-density images have even lower representation.
This is also supported by the improvements found from applying the tiled image augmentation, which artificially inflates the count of some training iterations.
\begin{table}
    \centering
    \fontsize{9}{9}\selectfont
    \setlength\tabcolsep{0.5em}
\begin{tabular}{ccccccccc}
 \toprule
 &\multicolumn{4}{c}{Val Set} & \multicolumn{4}{c}{Test Set}   \\
  \cmidrule(r){2-5} \cmidrule(r){6-9} 
 Limit & \# & \% & MAE & RMSE  & \# & \% & MAE & RMSE\\
\midrule
None & 0 & 0.00 &19.84  & 55.81 & 0 & 0.00 & 14.23  & 43.83\\
1000 & 3 & 0.31 & 17.94 & 43.99 &  1 & 0.09 & 12.96 & 26.69\\
500 & 12 & 1.26 & 16.11 & 34.80 & 7 & 0.66 & 12.12 & 23.02\\
\bottomrule
\end{tabular}

    \caption{
    \textbf{The effect of removing high-density images from FSC-133.}
    Excluding the small number of very dense images significantly improves our results, showing that these are a weakness of our method.
    \# and \% denote the total number and percentage of images excluded respectively.
    }
    \label{FSC147_exclusions_table}
\end{table}

\begin{figure}
    \centering
       \input{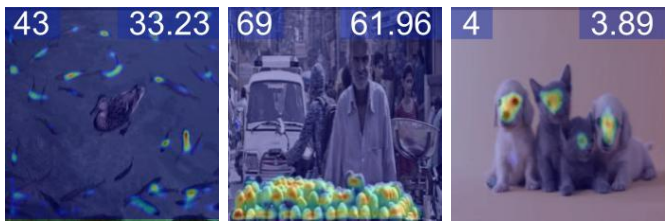}
    \caption{\textbf{Localisation of counting features with multiple classes present.} 
    The network either ignores all but one class (left and center) or it groups very similar classes and counts the superset (right). The left and center images have high levels of occlusion leading to low predictions.
    }
    \label{figure_multiclass}
\end{figure}

The main and most obvious limitation of RCC, and indeed any reference-less method, is that it is single-count. It finds the class most likely to be of interest and enumerates it, see Figure \ref{figure_multiclass}. Given the distribution of objects in FSC-147, this does not pose a problem during our evaluation.
Furthermore, the core achievement of this work is negating the requirement for test-time reference images and point-level annotations. There are also a wide range of applications to singular classes of novel objects, e.g.\ medical imaging \cite{xie2018microscopy}.
Nevertheless, adapting this method to generate multiple counts or a hierarchical class-count structure is scope for future work.

\vspace{5mm}
\subsection{Ablation Studies}
\label{results_ablationstudies}

\textbf{Attention-less features.}
To demonstrate that the global context provided by the attention mechanism present in vision transformers is a critical component to reference-less class-agnostic counting, we evaluate RCC with ResNet-50 \cite{he2016resnet} and ConvNeXt \cite{liu2022convnext} backbones in place of our transformer. 
We initialised both architectures with standard weights pre-trained on ImageNet \cite{deng2009imagenet}. 
While using weights generated by a supervised task that overlaps classes with our test and validation sets means that the results are not directly comparable to our self-supervised method, we believe that this comparison provides a best-case for these architectures.
Resnet-50 has a comparable accuracy to our vision transformer and ConvNeXt has a higher classification accuracy than our architecture when trained comparably \cite{liu2022convnext}.
To enable comparison across backbones, we use the same input feature size (224$\times$224) and take latent features at the same (28$\times$28) resolution as the patched features used in our backbone. 
As seen in Table \ref{ablation_arch}, even with the advantageous supervised pre-training, the attention-less features perform significantly worse than the self-supervised vision transformer.

\vspace{5mm}

\textbf{Count regression complexity.}
We assert that given sufficiently general and globally aware features, regressing an accurate count should be straightforward. To validate this, we compare our linear projection against two count regression heads. We found that more complex counting heads achieved equivalent or worse results. In Table \ref{ablation_arch}, we present results for a simple architecture, Conv(3$\times$3)-ReLU-Linear-Relu-Linear, and a more complex architecture comprised of four (3$\times$3) convolutional layers followed by three linear layers with ReLU activations. The convolutional heads perform worse than our projection on all three backbones. It appears that the extra computational capacity of the complex architecture is not only unnecessary for the task of counting but is in fact detrimental because it overfits to the training classes.
\begin{table}
    \centering
    \fontsize{9}{9}\selectfont
    \begin{tabular}{llrrrr}
\toprule
 & & \multicolumn{2}{c}{Val Set} & \multicolumn{2}{c}{Test Set} \\
 Backbone & Head & MAE & RMSE & MAE & RMSE \\
 \midrule

ResNet-50 & Projection & 31.80 & 78.87 & 24.99 & 75.87 \\
& Simple & 36.21 & 98.74 & 26.51 & 110.36\\
& Complex & 33.63 & 88.50 & 24.02 & 89.31
\\

ConvNeXt & Projection & 30.30 & 82.16 & 21.58 & 87.21 \\
 & Simple & 24.41 & 67.26 & 23.07 & 111.70 \\
 & Complex &  25.94 & 89.48 & 24.60 & 134.66 \\
ViT-Small & Projection & \textbf{19.84}  & \textbf{55.81} & \textbf{14.23}  & \textbf{43.83} \\

 & Simple & 23.08 & 66.19 & 17.46 & 78.60 \\
 &  Complex & 20.73 & 57.67 & 15.22 & 52.24 \\
\bottomrule
\end{tabular}

    \caption{\textbf{Performance of different feature backbones, and two more complex count regression head architectures on FSC-133.}
    }
    \label{ablation_arch}
\end{table}

\section{Conclusion}
\label{conclusion}
In this work, we present RCC, one of the first reference-less class-agnostic counting methods,
and show that it can be trained without point-level annotations.
This is based on the confirmed intuition that well-trained vision transformer features are both general enough and contextually aware enough to implicitly understand the underlying basis of counting, namely object detection and repetition identification. 
To evaluate and compare RCC against other methods, we use FSC-147, a standard counting dataset, and our proposed improved dataset FSC-133.
We demonstrate on both datasets that RCC is superior to the only other reference-less method and is competitive with current reference-based class-agnostic counting approaches that use full point-level supervision. We also show its cross-domain gerenalisability on CARPK and that it can localise appropriate object instances without any location-based intervention during training.
We believe that due to our lack of reliance on object class priors, reference images, and positional annotations, our method has significantly greater utility than other counting methods, especially when the composition or appearance of objects is uncertain.

Several extensions are possible for future research: \textit{first}, a multi-class output would be of clear utility as many real world applications will have a diverse range of objects; \textit{second}, a hierarchical system that is able to organise a multi-count output into likely groupings of objects could be useful to better understand the distribution of present types; \textit{third}, regressing an image of the type of object being counted could also aid in understanding the generated count.
\clearpage
\bibliographystyle{abbrvnat}
\bibliography{bib}

\begin{thebibliography}{50}
\providecommand{\natexlab}[1]{#1}
\providecommand{\url}[1]{\texttt{#1}}
\expandafter\ifx\csname urlstyle\endcsname\relax
  \providecommand{\doi}[1]{doi: #1}\else
  \providecommand{\doi}{doi: \begingroup \urlstyle{rm}\Url}\fi

\bibitem[Arteta et~al.(2014)Arteta, Lempitsky, Noble, and
  Zisserman]{arteta2014interactive}
C.~Arteta, V.~Lempitsky, J.~A. Noble, and A.~Zisserman.
\newblock Interactive object counting.
\newblock In \emph{European conference on computer vision}, pages 504--518.
  Springer, 2014.

\bibitem[Bai et~al.(2020)Bai, He, Qiao, Hu, Wu, and Yan]{bai2020adaptive}
S.~Bai, Z.~He, Y.~Qiao, H.~Hu, W.~Wu, and J.~Yan.
\newblock Adaptive dilated network with self-correction supervision for
  counting.
\newblock In \emph{Proceedings of the IEEE/CVF conference on computer vision
  and pattern recognition}, pages 4594--4603, 2020.

\bibitem[Barinova et~al.(2012)Barinova, Lempitsky, and
  Kholi]{barinova2012detection}
O.~Barinova, V.~Lempitsky, and P.~Kholi.
\newblock On detection of multiple object instances using hough transforms.
\newblock \emph{IEEE Transactions on Pattern Analysis and Machine
  Intelligence}, 34\penalty0 (9):\penalty0 1773--1784, 2012.

\bibitem[Borstel et~al.(2016)Borstel, Kandemir, Schmidt, Rao, Rajamani, and
  Hamprecht]{borstel2016gaussian}
M.~v. Borstel, M.~Kandemir, P.~Schmidt, M.~K. Rao, K.~Rajamani, and F.~A.
  Hamprecht.
\newblock Gaussian process density counting from weak supervision.
\newblock In \emph{European Conference on Computer Vision}, pages 365--380.
  Springer, 2016.

\bibitem[Cao et~al.(2018)Cao, Wang, Zhao, and Su]{cao2018scale}
X.~Cao, Z.~Wang, Y.~Zhao, and F.~Su.
\newblock Scale aggregation network for accurate and efficient crowd counting.
\newblock In \emph{Proceedings of the European conference on computer vision
  (ECCV)}, pages 734--750, 2018.

\bibitem[Carion et~al.(2020)Carion, Massa, Synnaeve, Usunier, Kirillov, and
  Zagoruyko]{carion2020DETR}
N.~Carion, F.~Massa, G.~Synnaeve, N.~Usunier, A.~Kirillov, and S.~Zagoruyko.
\newblock End-to-end object detection with transformers.
\newblock In \emph{European conference on computer vision}, pages 213--229.
  Springer, 2020.

\bibitem[Caron et~al.(2021)Caron, Touvron, Misra, J{\'e}gou, Mairal,
  Bojanowski, and Joulin]{caron2021DINO}
M.~Caron, H.~Touvron, I.~Misra, H.~J{\'e}gou, J.~Mairal, P.~Bojanowski, and
  A.~Joulin.
\newblock Emerging properties in self-supervised vision transformers.
\newblock In \emph{Proceedings of the IEEE/CVF International Conference on
  Computer Vision}, pages 9650--9660, 2021.

\bibitem[Chan and Vasconcelos(2009)]{chan2009bayesian}
A.~B. Chan and N.~Vasconcelos.
\newblock Bayesian poisson regression for crowd counting.
\newblock In \emph{2009 IEEE 12th international conference on computer vision},
  pages 545--551. IEEE, 2009.

\bibitem[Cholakkal et~al.(2020)Cholakkal, Sun, Khan, Khan, Shao, and
  Van~Gool]{cholakkal2020towards}
H.~Cholakkal, G.~Sun, S.~Khan, F.~S. Khan, L.~Shao, and L.~Van~Gool.
\newblock Towards partial supervision for generic object counting in natural
  scenes.
\newblock \emph{IEEE Transactions on Pattern Analysis and Machine
  Intelligence}, 2020.

\bibitem[Deng et~al.(2009)Deng, Dong, Socher, Li, Li, and
  Fei-Fei]{deng2009imagenet}
J.~Deng, W.~Dong, R.~Socher, L.-J. Li, K.~Li, and L.~Fei-Fei.
\newblock Imagenet: A large-scale hierarchical image database.
\newblock In \emph{2009 IEEE conference on computer vision and pattern
  recognition}, pages 248--255. Ieee, 2009.

\bibitem[Desai et~al.(2011)Desai, Ramanan, and
  Fowlkes]{desai2011discriminative}
C.~Desai, D.~Ramanan, and C.~C. Fowlkes.
\newblock Discriminative models for multi-class object layout.
\newblock \emph{International journal of computer vision}, 95\penalty0
  (1):\penalty0 1--12, 2011.

\bibitem[Do(2021)]{do2021attention}
P.~T. Do.
\newblock Attention in crowd counting using the transformer and density map to
  improve counting result.
\newblock In \emph{2021 8th NAFOSTED Conference on Information and Computer
  Science (NICS)}, pages 65--70. IEEE, 2021.

\bibitem[Dosovitskiy et~al.(2020)Dosovitskiy, Beyer, Kolesnikov, Weissenborn,
  Zhai, Unterthiner, Dehghani, Minderer, Heigold, Gelly,
  et~al.]{dosovitskiy2020vit}
A.~Dosovitskiy, L.~Beyer, A.~Kolesnikov, D.~Weissenborn, X.~Zhai,
  T.~Unterthiner, M.~Dehghani, M.~Minderer, G.~Heigold, S.~Gelly, et~al.
\newblock An image is worth 16x16 words: Transformers for image recognition at
  scale.
\newblock In \emph{International Conference on Learning Representations}, 2020.

\bibitem[Fan et~al.(2020)Fan, Zhuo, Tang, and Tai]{fan2020few}
Q.~Fan, W.~Zhuo, C.-K. Tang, and Y.-W. Tai.
\newblock Few-shot object detection with attention-rpn and multi-relation
  detector.
\newblock In \emph{Proceedings of the IEEE/CVF Conference on Computer Vision
  and Pattern Recognition}, pages 4013--4022, 2020.

\bibitem[Finn et~al.(2017)Finn, Abbeel, and Levine]{finn2017model}
C.~Finn, P.~Abbeel, and S.~Levine.
\newblock Model-agnostic meta-learning for fast adaptation of deep networks.
\newblock In \emph{International conference on machine learning}, pages
  1126--1135. PMLR, 2017.

\bibitem[Go et~al.(2021)Go, Byun, Park, Choi, Yoo, and Kim]{go2021fine}
H.~Go, J.~Byun, B.~Park, M.-A. Choi, S.~Yoo, and C.~Kim.
\newblock Fine-grained multi-class object counting.
\newblock In \emph{2021 IEEE International Conference on Image Processing
  (ICIP)}, pages 509--513. IEEE, 2021.

\bibitem[He et~al.(2016)He, Zhang, Ren, and Sun]{he2016resnet}
K.~He, X.~Zhang, S.~Ren, and J.~Sun.
\newblock Deep residual learning for image recognition.
\newblock In \emph{Proceedings of the IEEE conference on computer vision and
  pattern recognition}, pages 770--778, 2016.

\bibitem[Hsieh et~al.(2017)Hsieh, Lin, and Hsu]{hsieh2017drone}
M.-R. Hsieh, Y.-L. Lin, and W.~H. Hsu.
\newblock Drone-based object counting by spatially regularized regional
  proposal network.
\newblock In \emph{Proceedings of the IEEE international conference on computer
  vision}, pages 4145--4153, 2017.

\bibitem[Kang et~al.(2019)Kang, Liu, Wang, Yu, Feng, and Darrell]{kang2019few}
B.~Kang, Z.~Liu, X.~Wang, F.~Yu, J.~Feng, and T.~Darrell.
\newblock Few-shot object detection via feature reweighting.
\newblock In \emph{Proceedings of the IEEE/CVF International Conference on
  Computer Vision}, pages 8420--8429, 2019.

\bibitem[Lei et~al.(2021)Lei, Liu, Zhang, and Liu]{lei2021towards}
Y.~Lei, Y.~Liu, P.~Zhang, and L.~Liu.
\newblock Towards using count-level weak supervision for crowd counting.
\newblock \emph{Pattern Recognition}, 109:\penalty0 107616, 2021.

\bibitem[Li et~al.(2018)Li, Zhang, and Chen]{li2018csrnet}
Y.~Li, X.~Zhang, and D.~Chen.
\newblock Csrnet: Dilated convolutional neural networks for understanding the
  highly congested scenes.
\newblock In \emph{Proceedings of the IEEE conference on computer vision and
  pattern recognition}, pages 1091--1100, 2018.

\bibitem[Liang et~al.(2021)Liang, Chen, Xu, Zhou, and Bai]{liang2021transcrowd}
D.~Liang, X.~Chen, W.~Xu, Y.~Zhou, and X.~Bai.
\newblock Transcrowd: Weakly-supervised crowd counting with transformer.
\newblock \emph{arXiv preprint arXiv:2104.09116}, 2021.

\bibitem[Liang et~al.(2022)Liang, Xu, and Bai]{liang2022end}
D.~Liang, W.~Xu, and X.~Bai.
\newblock An end-to-end transformer model for crowd localization.
\newblock \emph{arXiv preprint arXiv:2202.13065}, 2022.

\bibitem[Lin et~al.(2021)Lin, Hong, and Wang]{lin2021object}
H.~Lin, X.~Hong, and Y.~Wang.
\newblock Object counting: You only need to look at one.
\newblock \emph{arXiv preprint arXiv:2112.05993}, 2021.

\bibitem[Liu et~al.(2018)Liu, Van De~Weijer, and Bagdanov]{liu2018leveraging}
X.~Liu, J.~Van De~Weijer, and A.~D. Bagdanov.
\newblock Leveraging unlabeled data for crowd counting by learning to rank.
\newblock In \emph{Proceedings of the IEEE conference on computer vision and
  pattern recognition}, pages 7661--7669, 2018.

\bibitem[Liu et~al.(2022)Liu, Mao, Wu, Feichtenhofer, Darrell, and
  Xie]{liu2022convnext}
Z.~Liu, H.~Mao, C.-Y. Wu, C.~Feichtenhofer, T.~Darrell, and S.~Xie.
\newblock A convnet for the 2020s.
\newblock In \emph{Proceedings of the IEEE/CVF Conference on Computer Vision
  and Pattern Recognition}, pages 11976--11986, 2022.

\bibitem[Lu et~al.(2018)Lu, Xie, and Zisserman]{Lu18}
E.~Lu, W.~Xie, and A.~Zisserman.
\newblock Class-agnostic counting.
\newblock In \emph{Asian Conference on Computer Vision}, 2018.

\bibitem[Marana et~al.(1997)Marana, Velastin, Costa, and
  Lotufo]{marana1997estimation}
A.~N. Marana, S.~Velastin, L.~Costa, and R.~Lotufo.
\newblock Estimation of crowd density using image processing.
\newblock \emph{Image Processing for Security Applications}, pages 1--8, 1997.

\bibitem[Michaelis et~al.(2018)Michaelis, Ustyuzhaninov, Bethge, and
  Ecker]{michaelis2018one}
C.~Michaelis, I.~Ustyuzhaninov, M.~Bethge, and A.~S. Ecker.
\newblock One-shot instance segmentation.
\newblock \emph{arXiv preprint arXiv:1811.11507}, 2018.

\bibitem[Mundhenk et~al.(2016)Mundhenk, Konjevod, Sakla, and
  Boakye]{mundhenk2016large}
T.~N. Mundhenk, G.~Konjevod, W.~A. Sakla, and K.~Boakye.
\newblock A large contextual dataset for classification, detection and counting
  of cars with deep learning.
\newblock In \emph{European conference on computer vision}, pages 785--800.
  Springer, 2016.

\bibitem[Ranjan and Hoai(2022)]{ranjan2022exemplar}
V.~Ranjan and M.~Hoai.
\newblock Exemplar free class agnostic counting.
\newblock \emph{arXiv preprint arXiv:2205.14212}, 2022.

\bibitem[Ranjan et~al.(2021)Ranjan, Sharma, Nguyen, and Hoai]{ranjan2021Famnet}
V.~Ranjan, U.~Sharma, T.~Nguyen, and M.~Hoai.
\newblock Learning to count everything.
\newblock In \emph{Proceedings of the IEEE/CVF Conference on Computer Vision
  and Pattern Recognition}, pages 3394--3403, 2021.

\bibitem[Redmon et~al.(2016)Redmon, Divvala, Girshick, and
  Farhadi]{redmon2016YOLO}
J.~Redmon, S.~Divvala, R.~Girshick, and A.~Farhadi.
\newblock You only look once: Unified, real-time object detection.
\newblock In \emph{Proceedings of the IEEE conference on computer vision and
  pattern recognition}, pages 779--788, 2016.

\bibitem[Ren et~al.(2015)Ren, He, Girshick, and Sun]{ren2015fasterrcnn}
S.~Ren, K.~He, R.~Girshick, and J.~Sun.
\newblock Faster r-cnn: Towards real-time object detection with region proposal
  networks.
\newblock \emph{Advances in neural information processing systems}, 28, 2015.

\bibitem[Sam et~al.(2019)Sam, Sajjan, Maurya, and Babu]{sam2019almost}
D.~B. Sam, N.~N. Sajjan, H.~Maurya, and R.~V. Babu.
\newblock Almost unsupervised learning for dense crowd counting.
\newblock In \emph{Proceedings of the AAAI Conference on Artificial
  Intelligence}, volume~33, pages 8868--8875, 2019.

\bibitem[Savner and Kanhangad(2022)]{savner2022crowdformer}
S.~S. Savner and V.~Kanhangad.
\newblock Crowdformer: Weakly-supervised crowd counting with improved
  generalizability.
\newblock \emph{arXiv preprint arXiv:2203.03768}, 2022.

\bibitem[Shi et~al.(2022)Shi, Lu, Feng, Liu, and Cao]{shi2022represent}
M.~Shi, H.~Lu, C.~Feng, C.~Liu, and Z.~Cao.
\newblock Represent, compare, and learn: A similarity-aware framework for
  class-agnostic counting.
\newblock In \emph{Proceedings of the IEEE/CVF Conference on Computer Vision
  and Pattern Recognition}, pages 9529--9538, 2022.

\bibitem[Sindagi and Patel(2019)]{sindagi2019ha}
V.~A. Sindagi and V.~M. Patel.
\newblock Ha-ccn: Hierarchical attention-based crowd counting network.
\newblock \emph{IEEE Transactions on Image Processing}, 29:\penalty0 323--335,
  2019.

\bibitem[Sokhandan et~al.(2020)Sokhandan, Kamousi, Posada, Alese, and
  Rostamzadeh]{sokhandan2020few}
N.~Sokhandan, P.~Kamousi, A.~Posada, E.~Alese, and N.~Rostamzadeh.
\newblock A few-shot sequential approach for object counting.
\newblock \emph{arXiv preprint arXiv:2007.01899}, 2020.

\bibitem[Sun et~al.(2021)Sun, Liu, Probst, Paudel, Popovic, and
  Van~Gool]{sun2021boosting}
G.~Sun, Y.~Liu, T.~Probst, D.~P. Paudel, N.~Popovic, and L.~Van~Gool.
\newblock Boosting crowd counting with transformers.
\newblock \emph{arXiv preprint arXiv:2105.10926}, 2021.

\bibitem[Touvron et~al.(2021)Touvron, Cord, Douze, Massa, Sablayrolles, and
  J{\'e}gou]{touvron2021training}
H.~Touvron, M.~Cord, M.~Douze, F.~Massa, A.~Sablayrolles, and H.~J{\'e}gou.
\newblock Training data-efficient image transformers \& distillation through
  attention.
\newblock In \emph{International Conference on Machine Learning}, pages
  10347--10357. PMLR, 2021.

\bibitem[Vaswani et~al.(2017)Vaswani, Shazeer, Parmar, Uszkoreit, Jones, Gomez,
  Kaiser, and Polosukhin]{vaswani2017attention}
A.~Vaswani, N.~Shazeer, N.~Parmar, J.~Uszkoreit, L.~Jones, A.~N. Gomez,
  {\L}.~Kaiser, and I.~Polosukhin.
\newblock Attention is all you need.
\newblock \emph{Advances in neural information processing systems}, 30, 2017.

\bibitem[Wang et~al.(2015)Wang, Zhang, Yang, Liu, and Cao]{wang2015deep}
C.~Wang, H.~Zhang, L.~Yang, S.~Liu, and X.~Cao.
\newblock Deep people counting in extremely dense crowds.
\newblock In \emph{Proceedings of the 23rd ACM international conference on
  Multimedia}, pages 1299--1302, 2015.

\bibitem[Wang et~al.(2022)Wang, Liu, Long, Sang, Xia, and Sang]{wang2022joint}
F.~Wang, K.~Liu, F.~Long, N.~Sang, X.~Xia, and J.~Sang.
\newblock Joint cnn and transformer network via weakly supervised learning for
  efficient crowd counting.
\newblock \emph{arXiv preprint arXiv:2203.06388}, 2022.

\bibitem[Wang and Wang(2011)]{wang2011automatic}
M.~Wang and X.~Wang.
\newblock Automatic adaptation of a generic pedestrian detector to a specific
  traffic scene.
\newblock In \emph{CVPR 2011}, pages 3401--3408. IEEE, 2011.

\bibitem[Xie et~al.(2018)Xie, Noble, and Zisserman]{xie2018microscopy}
W.~Xie, J.~A. Noble, and A.~Zisserman.
\newblock Microscopy cell counting and detection with fully convolutional
  regression networks.
\newblock \emph{Computer methods in biomechanics and biomedical engineering:
  Imaging \& Visualization}, 6\penalty0 (3):\penalty0 283--292, 2018.

\bibitem[Yang et~al.(2021)Yang, Su, Hsu, and Chen]{yang2021cfoc}
S.-D. Yang, H.-T. Su, W.~H. Hsu, and W.-C. Chen.
\newblock Class-agnostic few-shot object counting.
\newblock In \emph{Proceedings of the IEEE/CVF Winter Conference on
  Applications of Computer Vision}, pages 870--878, 2021.

\bibitem[Yang et~al.(2020)Yang, Li, Wu, Su, Huang, and Sebe]{yang2020weakly}
Y.~Yang, G.~Li, Z.~Wu, L.~Su, Q.~Huang, and N.~Sebe.
\newblock Weakly-supervised crowd counting learns from sorting rather than
  locations.
\newblock In \emph{European Conference on Computer Vision}, pages 1--17.
  Springer, 2020.

\bibitem[Zhang et~al.(2015)Zhang, Li, Wang, and Yang]{zhang2015cross}
C.~Zhang, H.~Li, X.~Wang, and X.~Yang.
\newblock Cross-scene crowd counting via deep convolutional neural networks.
\newblock In \emph{Proceedings of the IEEE conference on computer vision and
  pattern recognition}, pages 833--841, 2015.

\bibitem[Zhang et~al.(2016)Zhang, Zhou, Chen, Gao, and Ma]{zhang2016single}
Y.~Zhang, D.~Zhou, S.~Chen, S.~Gao, and Y.~Ma.
\newblock Single-image crowd counting via multi-column convolutional neural
  network.
\newblock In \emph{Proceedings of the IEEE conference on computer vision and
  pattern recognition}, pages 589--597, 2016.

\end{thebibliography}
\clearpage
\onecolumn
\appendices
\section{FSC-147 and FSC-133}
\subsection{Similar and Different Images}\label{similaranddifferent}
\begin{figure*}[!ht]
    \centering
    \fontsize{9}{9}\selectfont
    \input{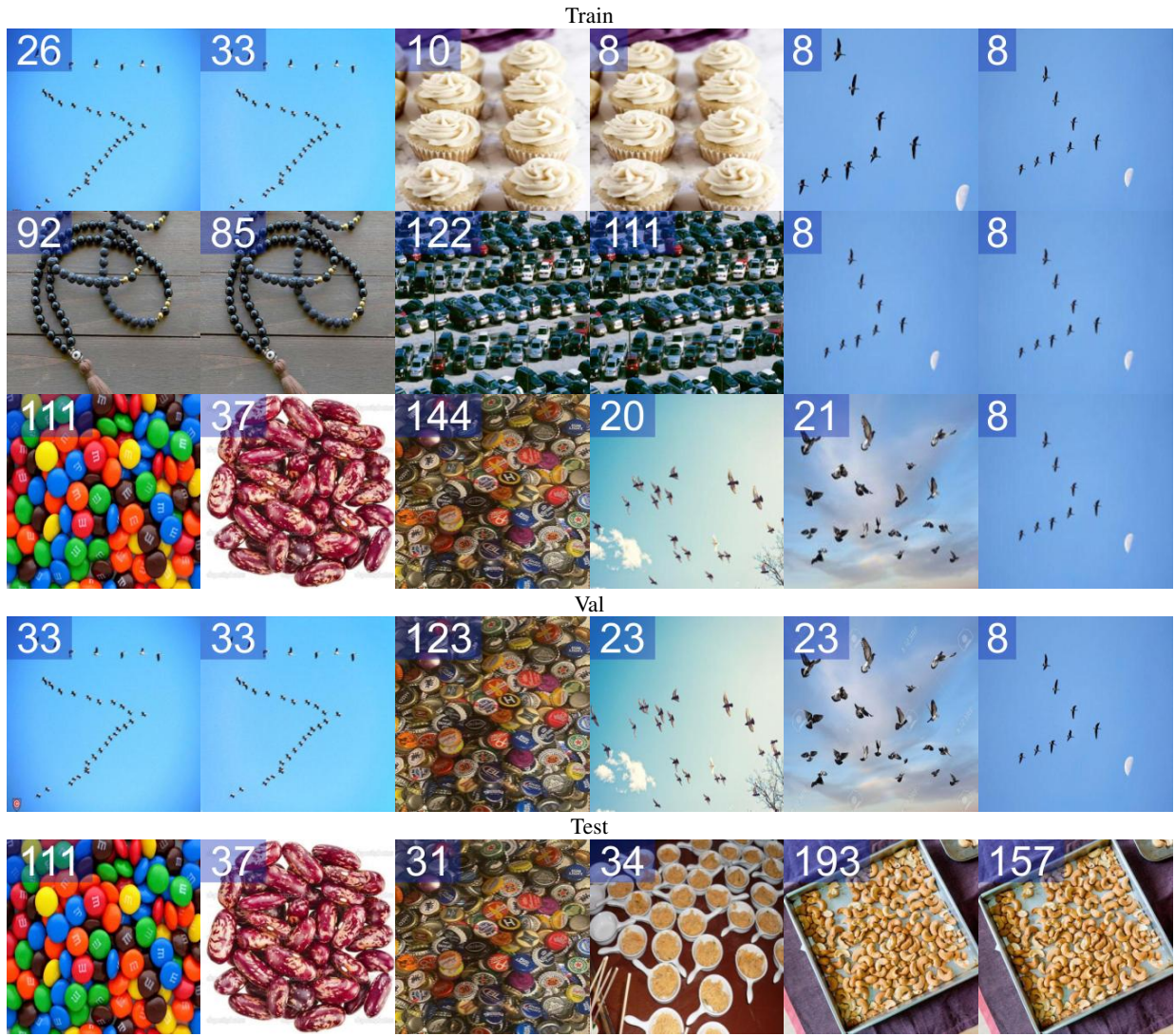}
    \caption{Examples of some of the 448 identical or close to identical images that appear in FSC-147 with different image IDs and their associated `ground truth' counts . Duplicates can occur with different count labels and/or in different splits.}
    \label{subset_dupes}
\end{figure*}

\begin{figure*}[!ht]   
    \centering
    \fontsize{9}{9}\selectfont
    \input{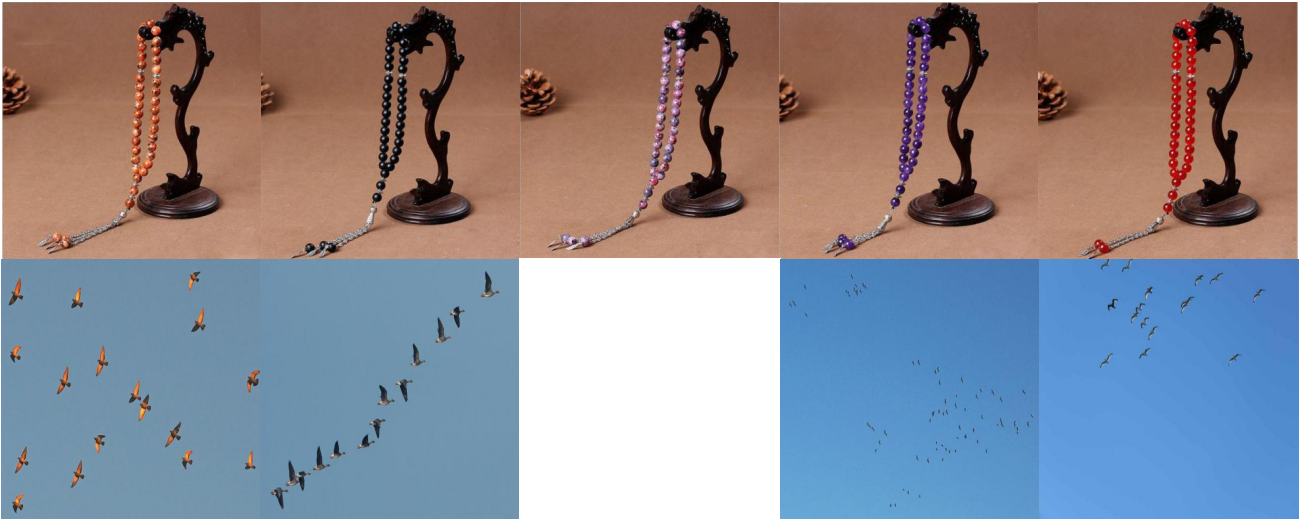}
    \caption{Images that have a low pixel-wise difference that we deemed to be different.}
    \label{no_similar}
\end{figure*}
\begin{figure*}[!ht]
    \centering
    \fontsize{9}{9}\selectfont
    \input{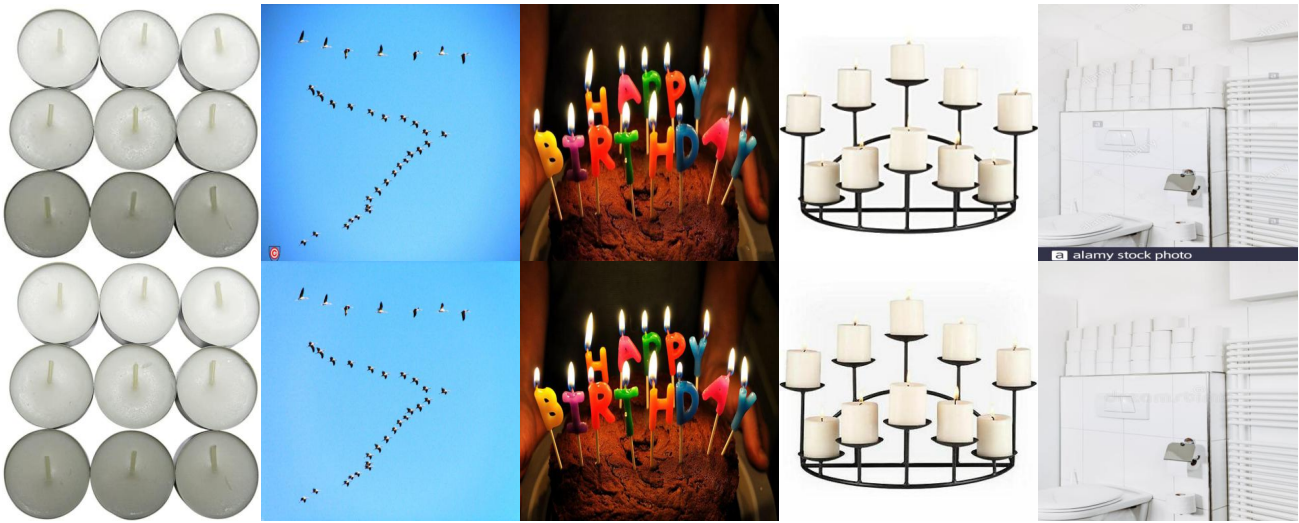}
    \caption{Examples of images at the upper bound of pixel-wise difference we considered that we deemed the same image. The pixel-wise differences of the images in each column when resized to 224$\times$224 are 5207, 5403, 7551, 8224 and 9396 respectively.}
    \label{similar}
\end{figure*}

\clearpage
\subsection{Similar Classes}\label{append_similar}
\begin{figure*}[!ht]    
    \centering
    \fontsize{8}{8}\selectfont
    \input{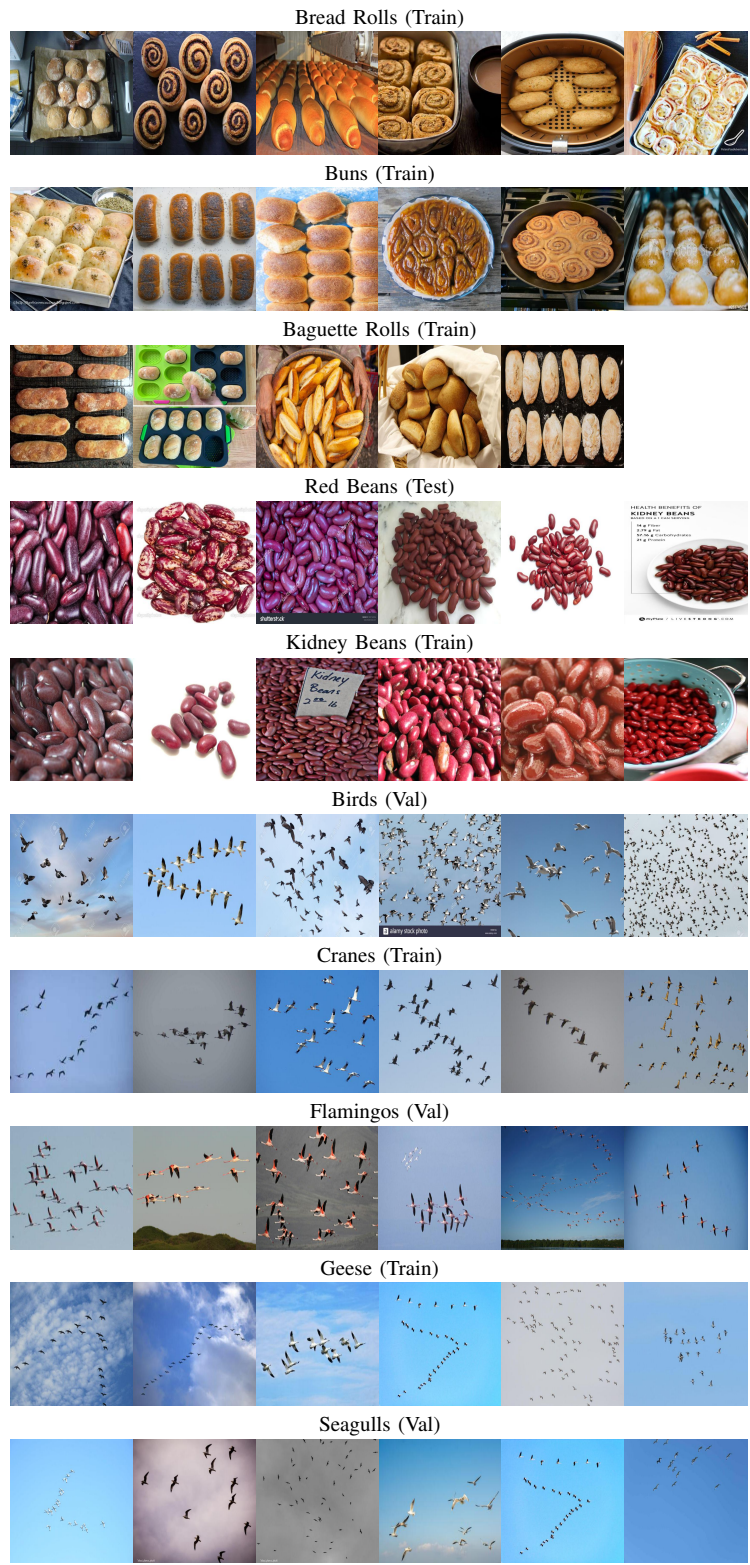}
    \caption{Examples of clear overlap or similarity between classes. Each row has a different class label in FSC-147. In FSC-133 these are simplified to Bread Rolls (Train), Kidney Beans (Train) and Birds (Train).}
    \label{arethesame}
\end{figure*}

\clearpage


\clearpage

\subsection{Duplicate Discrepancies}\label{discrepancies}
\begin{table*}[!ht]
    \centering
    \fontsize{9}{9}\selectfont
    \resizebox*{0.42\textheight}{!}{
        \begin{tabular}{clrlclrlrlc}
 \toprule
\multicolumn{4}{c}{Image A} & \multicolumn{4}{c}{Image B} & \multicolumn{2}{c}{Count Diff} \\
ID & Class & Count & Set & Id & Class & Count & Set & Abs & Rel & Kept\\
\cmidrule(r){1-4} \cmidrule(r){5-8} \cmidrule(r){9-10} \cmidrule(r){11-11}

4399 & cranes & 26 & train & 4549 & flamingos & 33 & val & 7 & 0.21 & A\\
4399 & cranes & 26 & train & 4719 & seagulls & 33 & val & 7 & 0.21 & A \\
2891 & caps & 144 & train & 1896 & bottle caps & 123 & val & 21 & 0.15 & B \\
4386 & cranes & 20 & train & 7415 & birds & 23 & val & 3 & 0.13 & A\\
6567 & pigeons & 21 & train & 929 & birds & 23 & val & 2 & 0.09 & A\\
4664 & geese & 11 & train & 6873 & birds & 11 & val & 0 & 0.00 & A\\
4613 & geese & 26 & train & 6714 & birds & 26 & val & 0 & 0.00 & B\\
4350 & cranes & 8 & train & 4704 & seagulls & 8 & val & 0 & 0.00 & 4683 \\
4350 & cranes & 8 & train & 4707 & seagulls & 8 & val & 0 & 0.00 & 4683\\
3506 & m\&m pieces & 111 & train & 3698 & candy pieces & 111 & test & 0 & 0.00 & A\\
3791 & kidney beans & 37 & train & 3494 & red beans & 37 & test & 0 & 0.00 & A\\

\bottomrule
\end{tabular}}
        \caption{The 17 cases where the same image appears in more than on of the: train set, validation or test set. The right column denotes which of these images, or if a third images appears in FSC-133.}
    \label{FSC_SPLIT_LEAK}
\end{table*}

\begin{table*}[!ht]
    \centering
    \fontsize{9}{9}\selectfont
    \resizebox*{0.42\textheight}{!}{
    \begin{tabular}{clrlclrlrlc}
 \toprule
\multicolumn{4}{c}{Image A} & \multicolumn{4}{c}{Image B} & \multicolumn{2}{c}{Count Diff} \\
ID & Class & Count & Set & Id & Class & Count & Set & Abs & Rel & Kept\\
\cmidrule(r){1-4} \cmidrule(r){5-8} \cmidrule(r){9-10} \cmidrule(r){11-11} 
5223 & donuts tray & 9 & val & 5664 & donuts tray & 12 & val & 3 & 0.25 & A\\
613 & seagulls & 7 & val & 623 & seagulls & 9 & val & 2 & 0.22 & B\\
4399 & cranes & 26 & train & 4549 & flamingos & 33 & val & 7 & 0.21 & 4683\\
4399 & cranes & 26 & train & 4620 & geese & 33 & train & 7 & 0.21 & 4683\\
4399 & cranes & 26 & train & 4719 & seagulls & 33 & val & 7 & 0.21 & 4683\\
3812 & cupcakes & 10 & train & 5238 & cupcakes & 8 & train & 2 & 0.20 & B \\
5142 & cashew nuts & 193 & test & 5801 & cashew nuts & 157 & test & 36 & 0.19 & B\\
4634 & geese & 18 & train & 6143 & geese & 15 & train & 3 & 0.17 & A\\
4634 & geese & 18 & train & 6669 & geese & 15 & train & 3 & 0.17 & A \\
6850 & cereals & 25 & train & 7337 & cereals & 21 & train & 4 & 0.16 & B\\
6839 & watches & 58 & test & 989 & watches & 69 & test & 11 & 0.16 & A\\
1896 & bottle caps & 123 & val & 2891 & caps & 144 & train & 21 & 0.15 & A\\
2717 & oranges & 94 & train & 6573 & oranges & 110 & train & 16 & 0.15 & A\\
3648 & cashew nuts & 35 & test & 5141 & cashew nuts & 30 & test & 5 & 0.14 & B\\
3014 & mini blinds & 26 & train & 7307 & mini blinds & 30 & train & 4 & 0.13 & A \\
4386 & cranes & 20 & train & 7415 & birds & 23 & val & 3 & 0.13 & A\\
2970 & fishes & 16 & train & 6953 & fishes & 18 & train & 2 & 0.11 & B\\
5929 & eggs & 76 & test & 6852 & eggs & 68 & test & 8 & 0.11 & B\\
4295 & green peas & 110 & test & 5504 & green peas & 122 & test & 12 & 0.10 & A\\
6567 & pigeons & 21 & train & 929 & birds & 23 & val & 2 & 0.09 & A\\
2743 & oranges & 57 & train & 7363 & oranges & 52 & train & 5 & 0.09 & B\\
6138 & cars & 122 & train & 6864 & cars & 111 & train & 11 & 0.09 & A\\
4220 & biscuits & 11 & train & 5422 & biscuits & 10 & train & 1 & 0.09 & B\\
3282 & finger foods & 31 & test & 3285 & finger foods & 34 & test & 3 & 0.09 & A\\
3462 & beads & 92 & train & 5638 & beads & 85 & train & 7 & 0.08 & A\\
4040 & beads & 117 & train & 5760 & beads & 108 & train & 9 & 0.08 & A\\
3492 & polka dots & 25 & val & 5738 & polka dots & 27 & val & 2 & 0.07 & A\\
3486 & polka dots & 81 & val & 5740 & polka dots & 87 & val & 6 & 0.07 & A\\
6917 & cereals & 29 & train & 7353 & cereals & 27 & train & 2 & 0.07 & B\\
5209 & donuts tray & 17 & val & 5656 & donuts tray & 16 & val & 1 & 0.06 & B\\
4013 & beads & 115 & train & 5754 & beads & 108 & train & 7 & 0.06 & B\\
524 & geese & 30 & train & 635 & cranes & 32 & train & 2 & 0.06 & B\\
4300 & green peas & 83 & test & 5506 & green peas & 88 & test & 5 & 0.06 & A\\
3287 & macarons & 16 & train & 428 & macarons & 17 & train & 1 & 0.06 & A \\
4634 & geese & 18 & train & 4675 & geese & 17 & train & 1 & 0.06 & A\\
6602 & pencils & 38 & train & 7400 & pencils & 36 & train & 2 & 0.05 & B \\
3475 & beads & 108 & train & 5637 & beads & 103 & train & 5 & 0.05 & A\\
2357 & bricks & 276 & train & 2401 & bricks & 263 & train & 13 & 0.05 & B\\
4044 & beads & 115 & train & 5335 & beads & 109 & train & 6 & 0.05 & A\\
4019 & beads & 105 & train & 5771 & beads & 109 & train & 4 & 0.04 & A\\
4049 & beads & 26 & train & 5755 & beads & 27 & train & 1 & 0.04 & B\\
3956 & candles & 46 & train & 5309 & candles & 44 & train & 2 & 0.04 & B \\
3969 & candles & 23 & train & 5317 & candles & 24 & train & 1 & 0.04 & A \\
3950 & candles & 24 & train & 5312 & candles & 25 & train & 1 & 0.04 & A\\
3020 & mini blinds & 34 & train & 7630 & mini blinds & 35 & train & 1 & 0.03 & A\\
3668 & toilet paper rolls & 31 & val & 5154 & toilet paper rolls & 32 & val & 1 & 0.03 & A\\
3679 & buns & 35 & train & 5026 & bread rolls & 36 & train & 1 & 0.03 & A\\
4014 & beads & 108 & train & 5749 & beads & 105 & train & 3 & 0.03 & A\\
3645 & cashew nuts & 60 & test & 5797 & cashew nuts & 58 & test & 2 & 0.03 & A\\
5148 & cashew nuts & 36 & test & 5803 & cashew nuts & 35 & test & 1 & 0.03 & A\\
7640 & apples & 32 & test & 7665 & apples & 33 & test & 1 & 0.03 & B\\
3819 & cupcakes & 36 & train & 5241 & cupcakes & 35 & train & 1 & 0.03 & B\\
3754 & pearls & 131 & train & 5185 & pearls & 127 & train & 4 & 0.03 & B\\
3666 & toilet paper rolls & 32 & val & 5157 & toilet paper rolls & 31 & val & 1 & 0.03 & B\\
3639 & jade stones & 31 & train & 5134 & jade stones & 32 & train & 1 & 0.03 & A\\
6200 & coins & 52 & train & 6228 & coins & 53 & train & 1 & 0.02 & A\\
267 & beads & 59 & train & 3242 & beads & 60 & train & 1 & 0.02 & A\\
6798 & birds & 87 & val & 976 & birds & 85 & val & 2 & 0.02 & B\\
3795 & kidney beans & 53 & train & 5547 & kidney beans & 52 & train & 1 & 0.02 & B\\
3955 & candles & 60 & train & 5306 & candles & 61 & train & 1 & 0.02 & A\\
2671 & bowls & 60 & train & 6724 & bowls & 59 & train & 1 & 0.02 & B\\
6738 & mini blinds & 53 & train & 7130 & mini blinds & 52 & train & 1 & 0.02 & B\\
5053 & beads & 139 & train & 5644 & beads & 136 & train & 3 & 0.02 & B\\
3426 & polka dots & 219 & val & 5046 & polka dots & 215 & val & 4 & 0.02 & B\\
3122 & coffee beans & 90 & train & 3134 & coffee beans & 89 & train & 1 & 0.01 & B\\
6961 & cartridges & 178 & train & 7680 & cartridges & 179 & train & 1 & 0.01 & A\\
4016 & beads & 109 & train & 5333 & beads & 108 & train & 1 & 0.01 & B\\
3469 & beads & 74 & train & 5050 & beads & 75 & train & 1 & 0.01 & A\\
4021 & beads & 106 & train & 5759 & beads & 107 & train & 1 & 0.01 & B\\
5149 & cashew nuts & 111 & test & 5807 & cashew nuts & 110 & test & 1 & 0.01 & B\\
4044 & beads & 115 & train & 5767 & beads & 116 & train & 1 & 0.01 & A\\
\bottomrule
\end{tabular}
}
     \caption{The 71 cases where the same image appears more than once in FSC-147 with different associated counts. The right column denotes which of these images, or if a third images appears in FSC-133.}
    \label{FSC_COUNT_DIFF}
    
\end{table*}
\clearpage

\subsection{FSC-147 and FSC-133 Class Split}\label{append_classes}
\begin{table*}[!ht]
    \centering
    \fontsize{9}{9}\selectfont
    \begin{tabular}{ccccc}
 \toprule

\multicolumn{5}{c}{Train} \\
\midrule
\textbf{alcohol bottles}&\textbf{baguette rolls}&balls&bananas&beads\\
bees&birthday candles&biscuits&boats&bottles\\
bowls&boxes&bread rolls&bricks&buffaloes\\
\textbf{buns}&calamari rings&candles&cans&caps\\
cars&cartridges&cassettes&cement bags&cereals\\
chewing gum pieces&chopstick&clams&coffee beans&coins\\
cotton balls&cows&\textbf{cranes}&crayons&croissants\\
\textbf{crows}&cupcake tray&cupcakes&cups&fishes\\
\textbf{geese}&gemstones&go game&goats&goldfish snack\\
ice cream&instant noodles&jade stones&jeans&kidney beans\\
kitchen towels&lighters&lipstick&m\&m pieces&macarons\\
matches&meat skewers&mini blinds&mosaic tiles&naan bread\\
nails&nuts&onion rings&oranges&pearls\\
pencils&penguins&pens&people&peppers\\
\textbf{pigeons}&plates&polka dot tiles&potatoes&rice bags\\
roof tiles&screws&shoes&spoon&spring rolls\\
stairs&stapler pins&straws&supermarket shelf&\textbf{swans}\\
tomatoes&watermelon&windows&zebras\\

\midrule
\multicolumn{5}{c}{Val} \\
\midrule
ants&\textcolor{blue}{birds}&books&bottle caps&bullets\\
camels&chairs&chicken wings&donuts tray&\textbf{\textcolor{blue}{flamingos}}\\
flower pots&flowers&fresh cut&grapes&horses\\
kiwis&milk cartons&oyster shells&oysters&peaches\\
pills&polka dots&prawn crackers&sausages&\textbf{\textcolor{blue}{seagulls}}\\
shallots&shirts&skateboard&toilet paper rolls\\

\midrule
\multicolumn{5}{c}{Test} \\
\midrule
apples&candy pieces&carrom board pieces&\textbf{\textcolor{blue}{cashew nuts}}&comic books\\
crab cakes&deers&eggs&elephants&finger foods\\
green peas&hot air balloons&keyboard keys&legos&marbles\\
\textbf{\textcolor{blue}{markers}}&nail polish&potato chips&\textbf{\textcolor{blue}{red beans}}&\textbf{\textcolor{blue}{sauce bottles}}\\
sea shells&sheep&skis&stamps&sticky notes\\
strawberries&sunglasses&tree logs&watches\\

\bottomrule
\end{tabular}
    \caption{The breakdown of classes in FSC-147. 
    Note there are identical categories eg. `red beans' and `kidney beans'; `baguette rolls', `buns' and `bread rolls' are semantically the same, there are also hierarchical categories e.g. `cranes', `geese', `pigeons', `seagulls', `crows', `swans' and `flamingos' could also be classified as also `birds'. 
    We show the classes that are combined in FSC-133 in \textbf{bold} and show the classes that are moved to the training set in FSC-133 in \textcolor{blue}{blue}.}
    \label{classes_old_table}
\end{table*}

\clearpage

\subsection{Sets of repeated Images}\label{list_identical}
\noindent
\textbf{Sets of Identical Images}
[1896, 2891], [19, 3092], [20, 3093], [22, 3094], [23, 3095], [26, 3097], [267, 3242], [269, 3255], [27, 3098], [9, 3084], [3282, 3285], [427, 3286], [428, 3287], [429, 3288], [3353, 5033, 5717], [3361, 5036], [3362, 5027], [3370, 5722], [3372, 5734], [3375, 5716], [3392, 5732], [3397, 5035], [3400, 3687], [3426, 5046], [3462, 5638], [3465, 5643], [3469, 5050], [3472, 5761], [3475, 5637], [3486, 5740], [3492, 5738], [3494, 3791], [3521, 5079], [3522, 5097], [3531, 5089], [3534, 5093], [3535, 5073], [3538, 5064], [3540, 5084], [3542, 5071], [3543, 5098], [3544, 5085], [3546, 5068], [3549, 5076], [3553, 5095], [3556, 3564, 5069], [3558, 5078], [3562, 5114], [3565, 5115], [3570, 5086, 5107], [3578, 5082, 5117], [3585, 5108], [3587, 5110], [3592, 5111], [3607, 5125], [3625, 5136], [3626, 5126], [3627, 5518], [3628, 5129], [3639, 5134, 5528], [3644, 5794], [3645, 5797], [3646, 5145], [3647, 5144], [3648, 5141], [3652, 5147, 5806], [3656, 5798], [3666, 5157], [3668, 5154], [3669, 5152], [3671, 5153], [3672, 5156], [3675, 5155], [3679, 5026], [3754, 5185], [3759, 5229], [3760, 5221], [3762, 5203], [3764, 5670], [3766, 5202], [3767, 5228, 5673], [3768, 5211, 5655], [3769, 5220], [3772, 5213, 5659], [3773, 5201, 5671], [3775, 5215], [3778, 5212, 5661], [3781, 5217], [3782, 5210], [3790, 5230, 5548], [3793, 5233], [3795, 5547], [3800, 5235, 5553], [3812, 5238], [3815, 5247], [3816, 4196, 5257], [3818, 5243], [3819, 5241], [3824, 5253], [3833, 5254], [3950, 5312], [3955, 5306], [3956, 5309], [3959, 5320], [3964, 5307], [3969, 5317], [3995, 5355], [4013, 5754], [4014, 5749], [4016, 5333], [4019, 5771], [4021, 5759], [4040, 5760], [4044, 5335, 5767], [4049, 5755], [4052, 5334], [4114, 5391], [4134, 5393], [4147, 5398], [4159, 5407], [4197, 5419], [4205, 5436], [4206, 5437], [4218, 5434], [4220, 5422], [4241, 5448], [4248, 5446], [4262, 5780], [4290, 5458], [4300, 5506], [4337, 4666], [4373, 4692], [4374, 4392], [4375, 4683, 4707], [4377, 4658], [4399, 4549], [4620, 4719], [5023, 5712], [5034, 5727], [5044, 5746], [5053, 5644], [5090, 5113], [5101, 5106], [5122, 5680], [5142, 5801], [5148, 5803], [5149, 5807], [5200, 5666], [5204, 5662], [5207, 5667], [5209, 5656], [5214, 5658], [524, 635], [5278, 5332], [5344, 5750], [5411, 5609], [5460, 5503], [6917, 7353], [7640, 7665]

\textbf{Sets Duplicates Images}
[3184, 3198], [3628, 5129], [3565, 5115], [3648, 5141], [4019, 5771], [3020, 7630], [3646, 5145], [6567, 929], [3362, 5027], [3800, 5235, 5553], [3666, 5157, 5151], [6554, 6566], [4238, 4243], [4049, 5755], [3638, 5525], [3587, 5110], [3286, 427], [3542, 5071], [3818, 5243], [4350, 4375, 4683, 4704, 4707, 6684], [6200, 6228], [3534, 5093], [3553, 5095], [267, 3242], [2895, 7441], [4534, 4570], [3947, 5318], [6798, 976], [3795, 5547], [3668, 5154], [2674, 6759], [3824, 5253], [5460, 5503], [3288, 429], [3679, 5026], [3543, 5098], [3955, 5306], [5209, 5656], [3647, 5144], [4014, 5749], [3522, 5097], [3556, 3564, 5069], [4044, 5335, 5767], [3762, 5203], [3759, 5229], [3675, 5155], [4262, 5780], [3778, 5212, 5661], [4013, 5754], [5207, 5667], [4241, 5448], [6602, 7400], [5452, 5779], [3535, 5073], [3781, 5217], [3956, 5309], [3014, 7307], [4248, 5446], [3392, 5732], [4664, 6873], [2743, 7363], [524, 635], [3531, 5089], [1896, 2891], [5200, 5666], [3969, 5317], [3492, 5738], [3816, 4196, 5257], [5090, 5113], [5142, 5801], [5278, 5332], [4206, 5437], [3375, 5716], [3549, 5076], [6850, 7337], [5344, 5750], [3949, 3960], [5101, 5106], [3122, 3134], [3084, 9], [3782, 5210], [4399, 4549, 4620, 4719], [3544, 5085], [4244, 5777], [5204, 5662], [3717, 5161], [4147, 5398], [4377, 4658], [2671, 6724], [6743, 7506], [3353, 5033, 5717], [3743, 5168], [6138, 6864], [3361, 5036], [3656, 5798], [2009, 2038], [6961, 7680], [4337, 4666], [3833, 5254], [4218, 5434], [3475, 5637], [3645, 5797], [6006, 6010], [4300, 5506], [3627, 5518], [3287, 428], [6280, 7538], [4016, 5333], [5223, 5664], [3570, 5086, 5107], [3626, 5126, 5519], [5148, 5803], [5411, 5609], [4373, 4692], [3486, 5740], [3793, 5233], [3546, 5068], [3370, 5722], [3767, 5228, 5673], [4220, 5422], [3773, 5201, 5671], [22, 3094], [6738, 7130], [3397, 5035], [3585, 5108], [2717, 6573], [2970, 6953], [2357, 2401], [3672, 5156], [5053, 5644], [5122, 5680], [3775, 5215], [26, 3097], [3540, 5084], [6198, 6226], [6834, 6870], [19, 3092], [4134, 5393], [3644, 5794], [27, 3098], [3815, 5247], [3760, 5221], [4114, 5391], [3959, 5320], [4197, 5419], [4386, 7415], [7640, 7665], [3639, 5134, 5528], [3768, 5211, 5655], [3462, 5638], [3812, 5238], [2289, 7172], [3671, 5153], [3995, 5355], [269, 3255], [5034, 5727], [1949, 6765], [3629, 5128], [3521, 5079], [4629, 547], [4290, 5458], [3372, 5734], [3625, 5136], [3950, 5312], [5023, 5712], [4159, 5407, 5604], [4295, 5504], [3469, 5050], [3607, 5125], [5929, 6852], [3426, 5046], [3578, 5082, 5117], [3506, 3698], [3669, 5152], [3554, 5092], [3494, 3791], [3769, 5220], [3764, 5670], [5044, 5746], [4374, 4392], [3558, 5078], [6839, 989], [3465, 5643], [3538, 5064], [4040, 5760], [23, 3095], [4613, 6714], [20, 3093], [4021, 5759], [4357, 4380], [3652, 5147, 5806], [3282, 3285], [4052, 5334], [5149, 5807], [3964, 5307], [3766, 5202], [5214, 5658], [6917, 7353], [613, 623], [3772, 5213, 5659], [3562, 5114], [3400, 3687], [3790, 5230, 5548], [4205, 5436], [3592, 5111], [3819, 5241], [3472, 5761], [4634, 4675, 6143, 6669], [3723, 5616], [3958, 5313], [3754, 5185]

\clearpage
\section{Tiling Augmentation Ablation.}\label{append_tile}
We validate the use of our image tiling augmentation by testing various frequencies and tiling configurations, as shown in Table \ref{ablation_tile2}. 
We found that introducing the denser (4$\times$4) tiling had less effect. 
At this density of tiling, each object instance is very small within the total image, so it is no longer interpretable, providing little meaningful training. 
\begin{table}[!h]
    \centering
    \fontsize{9}{9}\selectfont
    \begin{tabular}{lllll}
\toprule
  & \multicolumn{2}{c}{Val Set} & \multicolumn{2}{c}{Test Set} \\
 Frequency & MAE & RMSE & MAE & RMSE
 \\
 \midrule
 \multicolumn{5}{l}{(2$\times$2)} \\
75\% & 20.10 & 60.23 & 13.72 & \textbf{31.26} \\
50\% & \textbf{19.84}  & \textbf{55.81} & 14.23  & 43.83 \\
25\% & 20.95 & 64.44 & 14.98 & 47.08 \\
\; 0\% & 21.76 & 68.68 & 15.72 & 77.07 \\
\multicolumn{5}{l}{(4$\times$4) or (2$\times$2)} \\
75\% & 20.73 & 63.21 & \textbf{13.38} & 38.61\\
50\% & 20.00 & 65.08 & 15.44 & 37.47 \\
25\% & 20.51 & 63.73 & 15.71 & 42.77\\
\; 0\% & 21.76 & 68.68 & 15.72 & 77.07\\
\bottomrule
\end{tabular}

    \caption{Performance results for varying sizes and frequencies of our tiling augmentation on FSC-147. Images are tiled in either a (2$\times$2) grid, or in one of a (4$\times$4) or (2$\times$2) grid with equal probability.
    }
    \label{ablation_tile2}
\end{table}
\end{document}